# An Agent-based framework for cooperation in Supply Chain

Benaissa Ezzeddine
Université le Havre
Le Havre, France

Benabdelhafid Abdellatif
Université le Havre
Le Havre, France

Benaissa Mounir
Université Elmanar, OASIS
Tunis, Tunisia

**Abstract**
Supply Chain coordination has become a critical success factor for Supply Chain management (SCM) and effectively improving the performance of organizations in various industries. Companies are increasingly located at the intersection of one or more corporate networks which are designated by "Supply Chain". Managing this chain is mainly based on an 'information sharing' and redeployment activities between the various links that comprise it. Several attempts have been made by industrialists and researchers to educate policymakers about the gains to be made by the implementation of cooperative relationships. The approach presented in this paper here is among the works that aim to propose solutions related to information systems distributed Supply Chains to enable the different actors of the chain to improve their performance. We propose in particular solutions that focus on cooperation between actors in the Supply Chain.

## 1. Introduction

Several tools and techniques and methods of decision support are deployed and implemented between actors of the extended enterprise to meet the need for cooperation. It is through this work that fits the main contribution presented in this paper. We proposed solutions related to information systems and Supply Chains through an intelligent platform that allows different companies to have the ability to best meet customer demands. The main objective of our work is to develop a distributed architecture based on Multi-Agent Systems (MAS) and Semantic Web Services (SWS) for assistance in collaborative decision-making in the context of the extended enterprise. In the context of inter-firm collaboration in the Supply Chain, it is necessary to respect the autonomous decisions of each actor.

The MAS contribution to overall consistency of individual decisions in a business or shares a common interest and adopts a cooperative approach. In the same vein, a company brought to challenge an earlier decision in accordance with its partners, will have to renegotiate with them, in order to find a new compromise that satisfies all stakeholders. In this section we'll outline the main research field that affects our problem and that is mainly based on the paradigm agent. According to various studies [1], it would be Mark Fox had proposed the first to organize the Supply Chain as a network of intelligent agents. Distributed systems such as MAS allow the representation of each autonomous member of the corporate network [2]. According to Parunak [3], the network organizations of production and distribution have the same characteristics as agents: autonomy, social ability, reactivity and proactivity. Other projects are facing problems related to decision making in Supply Chains and have used agent-based technologies [4]. The literature shows that projects using the MAS in the Supply Chain confront essentially three main problems: modeling, design and management (control). Also, how to solve these problems also differs depending on the project: for example, the number and role of agents vary considerably.

The rest of the paper is organized as follows. Section 2 present MAS and SCM background. Section 3 discusses some related works on SCM Agent Systems. Section 4 proposes a framework and describes the modeling of the proposed system. Finally, section 5 provides the conclusion.

## 2. Background

Firms today increasingly consider Supply Chain management (SCM) to be a major vehicle to gain a competitive advantage in turbulent markets. While firms have traditionally acted as sole economic entities in the market, they have begun to form strategic alliances with other firms, integrating their business processes, and consolidating their resources. The advancement of Information Technology (IT) has allowed firms that participate in SCM to share information across organizational boundaries, bringing about substantial performance increases. For example, the collection of sales information at the point-of-sale and the sharing of that information via an Electronic Data Interchange (EDI) have lowered costs in the ordering processes.

Researchers have recently begun to focus on Multi-Agent based approaches to address collaboration and information sharing problems [5]. Supply Chain firms are typically modeled as software agents, which pursue their own goals under certain constraints. Prior research has addressed diverse aspects in the Supply Chain, but questions still remain concerning the best method of addressing and resolving collaboration and information sharing problems. Traditionally, scholars have focused on the problem domain in which supply and demand uncertainties are low. In this context, the best strategy is to implement ''efficient Supply Chains'' [6] by lowering costs. However, a Supply Chain may experience uncertainty both in a high supply and in a high demand.

In high supply uncertainty, the Supply Chain suffers an evolving supply process in which manufacturing technology is emerging, and the supply base is unstable. The traditional approaches may have limited applicability in this context, because calculating analytical solutions are prohibitive even impossible as these uncertainties create increased complexities, resulting in a model that becomes overly complicated [7].

This state of the art shows that the MAS has become popular for modeling and monitoring of Supply Chains which consist of various components/ identities like supplier, manufacturer, factories, warehouses, distributions agents etc [8]. The increasing complexity of the software systems has constantly led to the evolution of new programming paradigms [9] as the agent approach appears as an interesting technology to simulate and reproduce the collaborative behavior. The growth of Web Service and Agent paradigms has led to the emergence of numerous studies that suggest an agent and Web Service integration. According to the multiplicity of strategies targeting this integration [10], we can distinguish different approaches within this framework that the essential: agent and Web Service as the same concept. Agents and Web Services are the same entities. All services are Web Services and are provided by agents (the underlying program is an agent) [11], [12].

Agent and Web Service as distinct concepts. Agents are able to describe their services as Web Services and search /use of Web Services by making connections between Web Services standards (WSDL, SOAP) and agents (SD FIPA, FIPA ACL) [13], [14]. These approaches are often limited to agents FIPA standards-compliant or use an integration module or gateway. A difficult aspect of this work is to bridge the gap between the poor and Semantically synchronous interactions of Web Services and Semantically rich interactions and asynchronous agents.

By exploiting the architectures listed above, we can achieve a rapprochement-Agent Web Service in various directions and for specific reasons, the main ones as follows [15].

The use of MAS as an entity in mediating the functional model of Web Services. Mediation on several levels. For example, the authors of the work [16], offer MAS "Proxy" in order to locate Web Services while in other works such as [17] ADM planning and composition of Web Services are presented.

The use of Web Services as a framework for architectural and technological develop MAS accessible through the Web. In this type of topics we find applications of agents who offer their capabilities through Web Services. Here there are two categories which differ essentially in their design.

An integrated approach: they are Web Services developed on a model agent to perform complex tasks such as transaction management or business interactions [18]. And [19] proposes a set of communication primitives borrowed agent communication languages supporting dynamic interactions. A decoupled design: from a given a priori MAS, a layer based on Web Services capabilities makes the agent available over the Web either to other agents of MAS or traditional client applications.

In our case, agents have their specific tasks and Web Services deal with all the features highly stressed by the outside world (partners, customers and providers). This Web Service are composed of software components or computational functions which have a set of operations which define a specific functionality[20] In this case a role of mediation is provided by a specific agent that facilitates the tasks of discovery and selection of appropriate services by exploiting the Semantic aspect of Web Services.

We offer a platform for cooperation between different actors in an extended enterprise, including the case of a Supply Chain port. A special feature of this platform is that each player wishing to question the system and perform an operation, can be either a client requesting a service or services or a service provider, is played both roles simultaneously. Agents to represent these actors (agent and provider agent client) and other performing various tasks have been created at this level.

In order to accomplish a task requested by users of the platform, an inter-agent is necessary. These agents will communicate by exchanging messages based ontology. It is here that lies the other feature of our system, the use of Semantic Web Service and Multi-Agent Systems not only to create a domain ontology as in [21], but also We will use this technology to propose a negotiation protocol based ontology. The domain ontology used in our case of common vocabulary between the agents representing the core of the platform to reduce the conflict in the interpretation and subsequently communicate without ambiguity, and to describe the application domain in question using a set of concepts and relationships between them.

Unlike many researchers who have chosen the protocol of FIPA Contract-Net interaction protocol as to solve the problem of task allocation, we propose to set up negotiation protocol based ontology as rules of and inference to be within reach of the agents in order to present the set of rules governing the negotiation. Our system is processing and interpretation of the application and uses the services available in the environment to meet the needs of clients who query the system requesting a particular service.

Our system has an ontology that allows the discovery agent to apply reasoning. This ontology contains the knowledge that the agent has the form of rules and by which it can autonomously while processing the request. It could be useful for keeping query history and subsequently the response time to customer demand is much lower than that of an agent does not have this intelligent behaviour. We call our approach to ontology of Semantic Web Service to serve as a depot containing the Semantic description of Web Services that use the system and the client platform seeks to consume.

# 3. Related works on SCM agent system

In the past, various methods for handling logistics information in a Supply Chain were proposed. Some of them focused on using the techniques of artificial intelligence. Several related works on multiple agents and SCM are stated as follows.

[22] proposed a simulation based framework for developing customized Supply Chain models using a library of software components. Supply-Chain agents were designed and played different roles as retailers, manufacturers, and transporters for inventory control.

[23] also utilized the characteristics of Multi-Agent Systems to solve problems of Supply Chains.

[24] described a framework of negotiation-based Multi-Agents for SCM. In that framework, an order could be automated or semi-automated through a negotiation process between functional agents.

[25] proposed collaborative agent system architecture and an infrastructure for collaborative agent systems. It was a general architecture for an Internet-based Multi-Agent System and was very suitable for managing complex Supply Chains in large manufacturing enterprises.

[26] provided a re-configurable, multi-level, and agent-based architecture called MASCOT (Multi-Agent Supply Chain Coordination Tool) for coordinating Supply-Chain planning and scheduling. The MASCOT agent system possessed the following key functions: coordination, integration with heterogeneous planning and scheduling modules, mixed-initiative decision support, and re-configurability.

[27] developed a framework of agent-based electronic markets (e-markets) to simulate the dynamic transaction situations from subcontractors. It consisted of the following phases for implementation: multilateral negotiation protocol for e-markets, establishment of e-market ontology, building a prototype for Multi-Agent e-markets, verification, and evaluation.

[28] proposed an agent-based software system for assisting in the decision making of SCM and for efficient and effective usage of EDI. In particular, they developed a model integrated computing-based SCM modeling environment to allow domain experts creating models of software agents to simulate and control the actual on-line negotiation processes.

[29] discussed how the agents worked together and divided the techniques into six parts: agents in B2C (product brokering, merchant brokering, negotiation), negotiation protocols and strategies (auctions and negotiation, contract), agents in B2B (agents for SCM, hybrid agent solutions for SCM), special agent mediated techniques for SCM (mobile agent, evolutionary and data-mining for SCM), tool for implementation (communication language and protocols, agent communication language, agent developing tools and toolkits, mobile agent platforms), and security issues in an agent-mediated EC.

[30] developed a Supply-Chain Web Centric System, called the SC-Web-CS, which could provide different domain entities such as services, providers, transports, ordering, manufacturers, customers, distributors, retailers, and search. The system also considered some functions in implementation, including virtual clustering mechanism, communication platform across the system, interoperability, mobility, customer-centric service, XML/RDF encoding, and prototype implementation. Most of the above approaches considered multiple agents and simulation techniques in single companies. They did not use Semantics to allow for flexible information query with different linguistic terms. Only [27] used the XML/RDF techniques to assist agents in accessing databases to capture different data. They, however, only gave a rough introduction and did not describe how to use the Semantic web in SCM. Berners-Lee proposed the architecture of the Semantic web in 2001, where in the contents in documents could be stored in XML and RDF formats. The document structures could thus be easily represented and stored. There were modules for rules, logics, and other inference methods to support the Semantic induction. The ontology vocabulary also provided a flexible linguistic matching when different terms were used.

[31] thought that Web Services included techniques of Web ontology and Web agents. They used the Web ontology language (OWL) to describe knowledge resources that could allow the use of linguistic terms for Semantic query. The agent systems could solve problems of B2B and B2C EC in the Web-based transaction environment with machine-to-machine interaction.

In this paper, we thus refer to the architectures and concepts proposed by [32] and [31] to design an intelligent system with Semantic web and multiple agents in a fashion industry SCM and to overcome the above problems.

[33] thought that the inter-company data exchange, procurement, and coordination of production in mass customization could be improved by means of a Multi-Agent System for SCM.

[34] presented a Web-based knowledge management system for facilitating seamless sharing of product data among application systems in virtual enterprise. The sharing of product data included metadata, ontology, mapping relationships and applications.

[35] then formulated agent strategies in a Supply Chain model based on the virtual market concept with multiple agents, and demonstrated the applicability of economic analysis to the framework by simulation under a dynamic environment.

[36] also indicated some key points on how agents worked in e-marketplaces of EC. One point was that collaborative agents working together could achieve the task of a knowledge sharing process. They built a framework of an e-Supply Chain to provide an intelligent e-Marketplace with multiple agents. In their framework, agents played different roles and took different responsibilities in the management process. These agents

included a buyer agent, a provider agent, a discovery agent, a transaction agent, and a monitoring agent.

[37] also proposed a negotiation methodology based on the Multi-Agent System for heterarchical and complex manufacturing control system.

## 4. Intelligent System for Extended Enterprise Cooperation (i-SEEC)

We covet to propose solutions related to information systems and Supply Chains through an intelligent platform (i-SEEC) that allows different companies to have the ability to best meet customer demands. The main objective of our work is to develop a distributed architecture based on Multi-Agent Systems (MAS) and Semantic Web Services (SWS) for assistance in collaborative decision-making in the context of the extended enterprise.

The proposed architecture allows:

- Decision-making by integrating distributed mechanisms for identifying dynamic services offered by various partners through cooperation and collaboration the various information systems.
- Provide interoperable collaboration and cooperation between partners regardless of any distinctive characteristics of certain players (technologies, norms and standards, platforms and software, etc. ....).
- Managing Semantic heterogeneity illustrated by the differences between concepts and vocabularies handled by the various stakeholders in ensuring a good understanding and proper interpretation of traded services.
- The timing of decisions by different actors to overcome the problems of information flow and product flow; interact and negotiate in order to propose solutions to various stakeholders while ensuring the autonomy and confidentiality of critical data in the case of independent actors or competitors.
- The integration of new players to cooperate and collaborate with stakeholders in the Supply Chain.

The Semantic Web Services provide a very high level of interworking. Nevertheless, they do not have enough degree of autonomy nor capacity to adapt in a dynamic way to the changing situations. In such situations, the intelligent agents can contribute to give the systems a high degree of autonomy and dynamicity. In conformity with this, we propose an architecture which intelligent Agents and the SWS work together in a co-operative way in a common environment by the means of ontology for the co-operation of the various links of the extended Enterprise. Actor's wishes to integrate i-SEEC will have the possibility of having two different interfaces, the interface consumer and the interface provider. The consumer makes the request of one or more services and the provider makes available to service consumers without full disclosure of all information. To do this, we make the description of various components of the platform i-SEEC (Fig. 1).

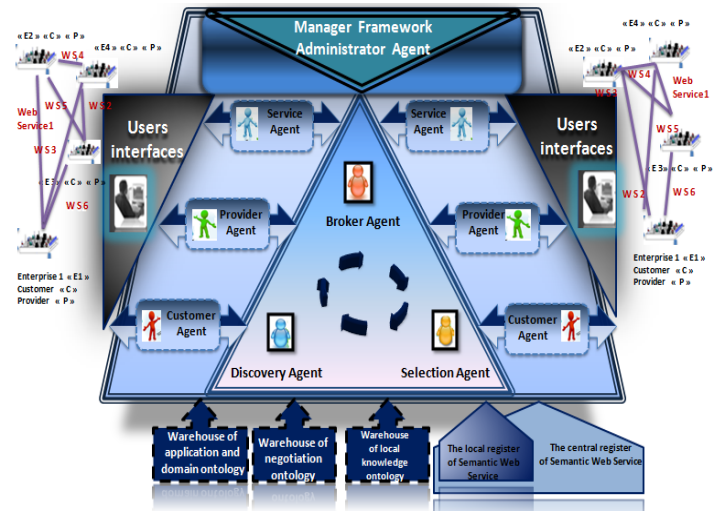

Fig. 1 i-SEEC Architecture.

4.1 Agents of the platform i-SEEC

**Administrator agent:** It plays a crucial role: the administrator agent deals with the management of the accounts of the users who wish integrated into the system in order to achieve a common goal. This agent ensures coherence between the contained information's in the local register of Semantic Web Services and those contained in the central register annotated semantically, in order to take into account the modifications carried out with an urgent data. Once an actor of the Supply Chain is added to the platform and his authentification is carried out. The administrator agent carries out the instanciation of the customer agent which designates the assistant of the customer, the provider agent representing the assistant of the provider and the service agent which will be detailed in the following part.

**The customer agent:** The customer agent allows the entry of the requests of the users system. He plays a central role for the initiation of discovered services while transmitting to the discovery agent the request of the customer. He deals with also the posting of the result after the treatment of this request. The internal architecture of the customer agent is illustrated in the following figure Fig. 2.

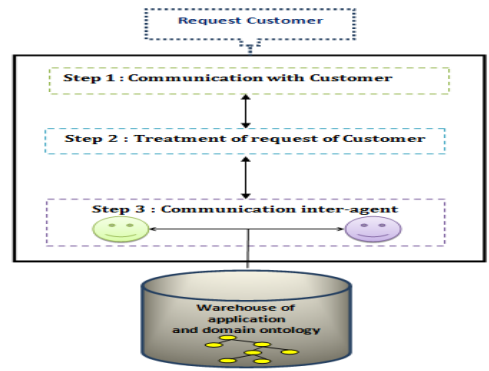

Fig. 2 Internal architecture of the customer agent.

**The provider agent:** It is the representative of the provider; it is used to be interfaced with the platform i-SEEC and the provider of the Web Service. It allows the recording of the Web Services described semantically in the database of the provider and the publication of the services in order to be used by the various links of the Supply Chain. It ensures also the update of the information connected to the Web Service.

**The service agent:** They are like representative abundant services by the provider. It intervenes at the time of the negotiation with the customer agent in order to draw up a contract, like giving further information more on the details of the service published by the provider in question. The agents described after represent lee core of our platform. They start at the time of its launching.

4.2 The core of the platform i-SEEC

The core of our platform is composed of three agents which start at the time of its launching:

**The discovery Agent:** The discovery agent deals with the request of the agent customer, where it carries out the research of the Semantic services Web appropriate to the customer requirements. The internal architecture of the discovery agent is illustrated in the following figure Fig.3.

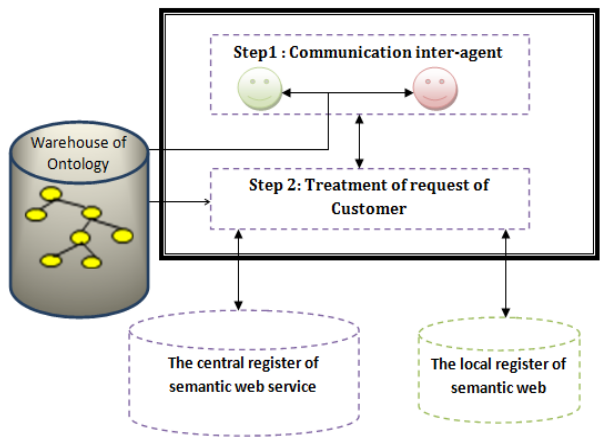

Fig. 3 Internal architecture of the customer agent.

For the step of communication inter-agent: the discovery agent receives the request of the customer agent. Then, it carries out the request handling. Once the treatment of the request is finished, it receives messages which constitute the answers at the request of the customer agent. These answers can be extracted the local register of service and if it service were not to find. An interrogation of the central register of services will take place then it service will be sent to the discovery agent.

Step 2: Treatment of request: at this stage two functions are assured.
The first function makes it possible to select the ontology of the field in question of the deposit of ontology of field and application. The second function corresponds to a mechanism of correspondence which will take place to meet the customer requirement.

**The selection agent:** The selection agent is responsible for the research of the best whole of service to satisfy the customer requirements for this it acts as follows.

- The selection agent makes a call of proposal has all the agents of service. Each service agent makes its proposal according to the preferences of the service provider.

- It chooses the best whole of services according to their utility envisaged in agreement with the preferences of the customer.

- The selection agent returns to the customer agent the overall matched list of services with the conditions agreed upon for the execution of each service.

- The customer agent shows the customer the list sorted services which selects that it would prefer and asks for the agent selection of put it in contact with the representative of which a negotiation process will take place in order to draw up the contract.

- The customer agent and the suitable service agents act one on the other so that they call really it selected services. During the invocation, the service agent must trace the information of entry of the customer with the parameters of entry of the methods to be carried out. This task is facilitated by employing the ontology of field in the representation of goal and the Semantic description of the services. The service agent returns the results of services to the agent customer, which responsible for is posted with the customer. The internal architecture of the selection agent is illustrated in the following figure.

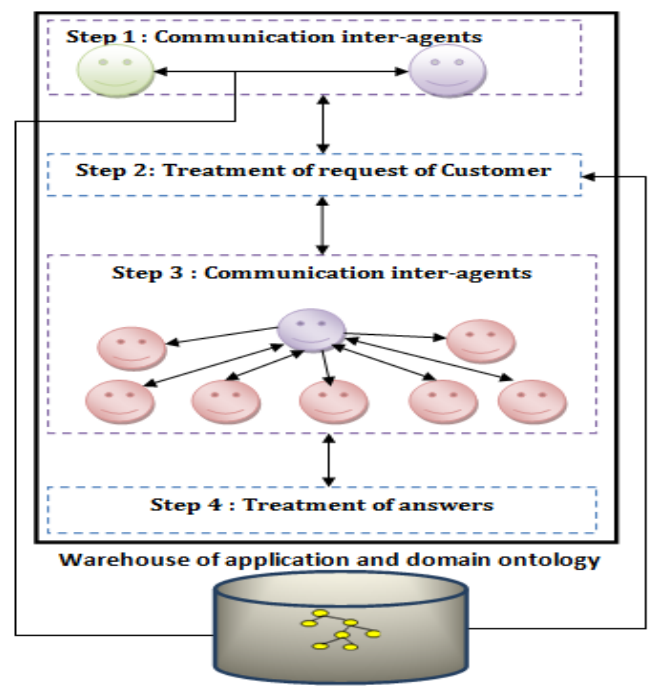

Fig. 4 Internal architecture of the selection agent.

**The broker agent:** It deals with the balance of the work, the monitoring of the execution process of the service. It is responsible to manage interoperability. However each agent of the platform can deal with the Semantic management of interoperability, and if the agent can't achieve the goal, he delegates the task to the broker agent.

We pass to present the other components of our platform: the warehouses of ontologies and the registers of Semantic Web Service. There exist three warehouses of ontologies:

**Warehouse of application and domain ontology:**

- The domain ontology: it represents a conceptualization of the specific field, while enumerating the terms which allow the description of this field and the relations between them. It allows the creation of a common vocabulary between the various agents without mistakes in interpretation.

- The application ontology: it includes concepts depending on a field and a particular task, which belong to the concepts of these two ontologies (ontology of field and ontology of task). With the execution of certain activities by the entities of the field, these concepts are often the parts played by these entities of the field [38].

**Warehouse of negotiation ontology:** This ontology includes the concepts and the rules used for the description of the protocol of negotiation based ontology which the agents are likely to use it to negotiate and establish an agreement on the conditions under which the services will be carried out to draw up a contract between them. We explain in details the protocol of negotiation used within the system i-SEEC.

**Warehouse of local knowledge ontology:** It contains for each agent the knowledge about the environment which it has. This ontology contains the knowledge about tasks affected for each agent as well as the mechanisms and the available resources in order to carry out these tasks.

As far as the registers of Semantic Web Service, there exist too types: central and local.

**The central register of Semantic Web Service:** It contains the Semantic description of the various abundant services by the various actors of the Supply Chain distributed. Each assistant of the provider deals with the control of the central register of Semantic Web Services. This register has an interface of communication which makes it possible the agent discovery to discover the services allowing the realization of an operation required by the customer and to the actors of the Supply Chain to publish the services that they wish to make public. Thanks to the service of discovery, the discovery agent can seek the service in question like their providers. The service of publication makes it possible to the providers to carry out the publication and the description of their services. The database of Semantics Web Services thus allows to the information storage concerning the abundant services their providers.

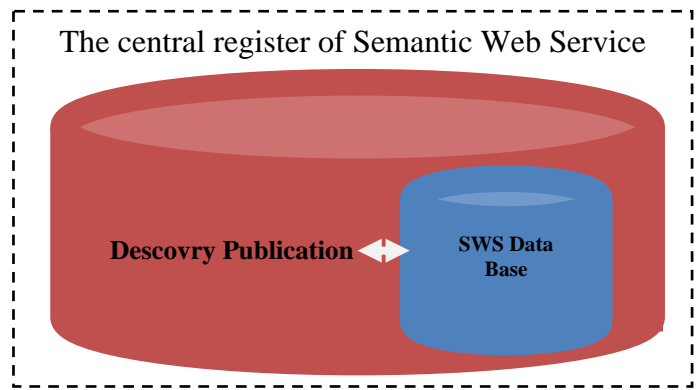

Fig. 5 The central register of Semantic Web Services

**The local register of Semantic Web Service:** It represents a copy of the central register of services of the providers. It allows the storage of the services most requested by the customers and the providers. It contributes to the minimization of the time discovered of the services. It also makes it possible to improve the relation between the customer and the provider with whom it has already to carry out the co-operation.

## 5. Modeling of i-SEEC

Modeling enables us to obtain a software prototype/quality and facilitates the comprehension of the operation of our platform. We present the digraphs of sequence to assemble the various interactions carried out between the various actors of the platform.

5.1 Management of the users of the platform i-SEEC

The management of the users (customer/ provider) of the platform is a task managed by the agent administrator of the platform which carries out the addition, as well as the update of information concerning the actors of the Supply Chain.

The sequence diagram illustrates the process of sequence of exchanges of the messages if a customer wishes to be integrated into the platform i-SEEC.

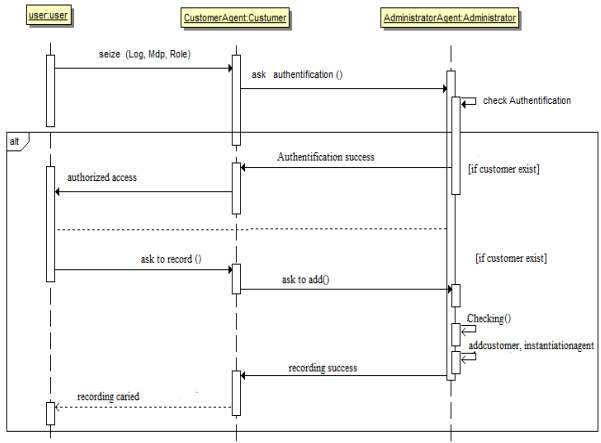

Fig. 6 The sequence diagram of management of the users of the platform i-SEEC.

If an actor of the Supply Chain wishes to reach his workspace. He authenticates himself near the system by seizing his login, password and his role (provider or customer). In this diagram we detail the mechanism of operation for the case where the actor of the Supply Chain connects himself as being customer. The customer agent transmits the request for connection to the agent administrator. This last carries out the checking of the data seized by the customer. Two cases of figure are presented during the checking if the customer exists and the seized data are coherent thus it sends in its turn the message of access authorization to the customer agent which carries out the posting of the result to the customer. In the contrary case a message will be returned to the customer agent which the seized data are erroneous. The customer will be informed by his assistant (customer agent) in order to obtain the good information. If the customer is not recorded yet within the system, he asks his recording. Once its addition is carried out, the agent administrator him instancie the agent which represents it within the system. It is the same mechanism as for the provider.

5.2 Discovered of a service

The objective of discovered services is to allow the customers and the providers wishing to carry out certain operations to identify the services which allow the realization of the activities concerned. This discovery is carried out by the discovery agent. The customer agent addresses a request to discover a service with the discovery agent. This last carries out the search for services required in the local register of Semantic Web Service in order to find to it services more often requested by the customers or the providers.
If the required services are not found in the local register of Semantic Web Services. The discovery agent carries out research in the central register of Semantic Web Services. The discovery agent carries out the invocation of the service of discovered central register of Semantic Web Service. The results of the discovery, local or central are transmitted then to the assistant of the customer (agent customer) to post them in all transparency with the final customer. The process of the discovery requires a Semantic mechanism of bringing together in order to establish the correspondences between the services sought by the customer and those defined by the services published in the central register of Semantic Web Services .

Following the discovery of the services, the discovery agent deals with the food of the local register with the whole of the services selected as well as the bond towards their providers.

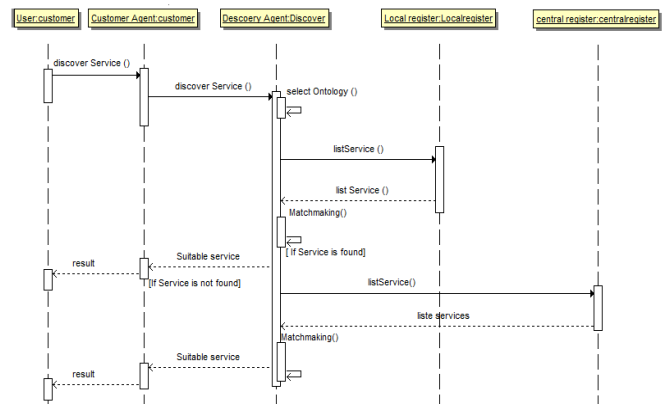

Fig. 7 The sequence diagram of discovered of a service.

## 6. Conclusion

Regarding the end-to-end communication, the system was based entirely on the prevailing and emerging Internet standards, ensuring platform, vendor and language independence. State-of-the-art technologies, such as HTTP, XML, SOAP and WSDL guarantee that the system is really open and future-proof. The use of a Service-based architecture keeps up with the evolution of Internet that is gradually transforming from a human-oriented to a service-oriented universal network, by extended use of Web Services. Totally XML-based information exchange between all the participant entities of Supply Chain provides a viable framework for universal connectivity without technology barriers, which could be imposed by closed and proprietary technologies. The proposed topology addresses many issues where future research could yield interesting results.

## References

[1] C. Dodd and S. Kumara, "A distributed model for value nets", IEA/ AIE, 2001, page 718-727.

[2] S. Chehbi, R. Derrouiche, Y. Ouzrout and A. Bouras, " Multi- Agent Supply Chain Architecture to Optimize Distributed Decision Making ", Proceedings of the 7th World Multiconference on Systemics, Cybernetics and informatics,SCI, 16, Orlando, USA, 2003.

[3] H.V.D Parunak., S. Brueckner, J. Sauter and R. Matthews, "Distinguishing Control and Plant Dynamics in Enterprise Modeling", Proceedings of the 2nd DARPA-JFACC Symposium on Advances in Enterprise Control. Minneapolis, USA, 2000.

[4] T. Moyaux, B. Chaib-draa, S. D'Amours, "Multi-agent coordination based on tokens: Reduction of the bullwhip effect in a forest Supply Chain", Proceedings of the 2nd International Joint Conference on Autonomous Agents and Multi- Agent Systems (AAMAS), Australie, 2003.

[5] J.M. Swaminathan, S.F. Smith and N.M. Sadeh, "Modeling Supply Chain dynamics: A multiagent approach", Decision Sciences 29 (30), 1998, 607-632.

[6] H. L. Lee, K. C. So and C.S. Tang, "The value of information sharing in a two-level Supply Chain", Management Science, 46(5), 2000, 626–643.

[7] J.F. Shapiro, "Modeling the Supply Chain", Pacific Grove, CA: Thompson Learning, 2001.

[8] V. Kumar and S. Srinivasan, "A Review of Supply Chain Management using Multi-Agent System", IJCSI International Journal of Computer Science Issues, Vol. 7, Issue 5, 2010.

[9] P. Jain and D. Dahiya, "An Architecture of a Multi Agent Enterprise Knowledge Management System Based on Service Oriented Architecture", IJCSI International Journal of Computer Science Issues, Vol. 9, Issue 2, No 3, 2012.

[10] J. Clément, J. Pascal and A. Stefano, "Intégration orientée service des modèles Grid et Multi-Agents", 14èmes Journées Francophones sur les Systèmes Multi-Agents, JFSMA06, 2006, pp. 271-274.

[11] F. Ishikawa, N. Yoshioka and Y. Tahara, "Toward Synthesis of Web Services and Mobile Agents", 2nd International Workshop on Web Services and Agent Based Engineering, WSABE'04, 2004, pp 48-55.

[12] J. Peters, "Integration of Mobile Agents and Web Services", 1st European Young Researchers Workshop on Service Oriented Computing (YR-SOC'05), Leicester, UK, Software Technology Research Laboratory, De Montfort University, 2005, pp 53-58.

[13] D. Greenwood and M. Calisti, "Engineering Web Service - Agent Integration", IEEE Systems, Cybernetics and Man Conference, The Hague, Netherlands, Washington, IEEE Computer Society, 2004.

[14] A. Seghrouchni and A. Suna, "Computational Language for Autonomous, Intelligent and Mobile Agent", Proceedings of the Programming MAS (PROMAS 2003) (Melbourne, Australia, July 15, 2003), Lecture Notes in Artificial Intelligence, vol. 3067, Springer-Verlag, Berlin Heidelberg (2004), 2004, pp. 90-110.

[15] E. Amal, H. Serge, M. Tarak and S. Alexandru, "Interopérabilité des Systèmes Multi-Agent en utilisant des Services Web", BOISSIER O., GUESSOUM Z., Eds., 12th Journées Francophones sur les Systèmes Multi-Agents, Paris, France, 2004.

[16] S.K, Mostefaoui and G. K, Mostefaoui, "Towards A Contextualisation of Service Discovery and Composition for Pervasive Environments", Workshop on Web services And Agent-based ingineering, Melbourne, Australia, 2003.

[17] D. Richards, S. Van Splunter, F. M. Brazier and M. Sabou, "Composing Web Services using an Agent Factory Workshop on Web services And Agent-based ingineering", Melbourne Australia, 2003.

[18] T. Jin, and S. Goschnick, "Utilizing Web Services in an Agent Based Transaction Model (ABT) ", Workshop on Web services And Agent-based ingineering, Melbourne, Australia, 2003.

[19] G. Petrone, "Managing flexible interaction with Web Services", Workshop on Web services And Agent-based engineering, Melbourne, Australia, 2003.

[20] B. B.Morales, X. M. Pasco and M. V. Lombardo, Survey: "Grid Computing and Semantic Web", IJCSI International Journal of Computer Science Issues, Vol. 7, Issue 3, No 5, 2010.

[21] S. Wei, P. Tzung and J. Rong, "A framework of E-SCM multi-agent systems in the fashion industry", Int. J. Production Economics, 114, 2008, 594–614.

[22] J. M. Swaminathan, S.F. Smith and N.M. Sadeh, "Modeling Supply Chain Dynamics: A Multiagent Approach", Decision Sciences, Volume 29 Number 3, 1998, pp. 607-632.

[23] W.E. Walsh and M.P. Wellman, "Modeling Supply Chain formation in multiagent systems", In: Agent Mediated Electronic Commerce (IJCAI Workshop), 1999, 94–101.

[24] Y. Chen, Y. Peng, T. Finin, Y. Labrou, S. Cost, B. Chu, J. Yao, R. Sun and B Wilhelm, "A negotiation based multi agent system for Supply Chain", In: Working Notes of the Agents '99 Workshop on Agents for Electronic Commerce and Managing the Internet-Enabled Supply Chain, Seattle,WA, April 1999.

[25] W. Shen, M. Ulieru, D. Norrie and R. Kremer, "Implementing the internet enabled Supply Chain through a collaborative agent system", In: Proceedings of Agents '99 Workshop on Agent Based Decision-Support for Managing the Internet- Enabled Supply-Chain, Seattle, 1999, pp. 55–62.

[26] N. Sadeh, D.W. Hildum, D. Kjenstad, and A. Tseng, "MASCOT: an agent-based architecture for dynamic Supply Chain creation and coordination in the internet economy", Prod Plan Control, vol. 12(3), 1999, pp: 212–223.

[27] K. Kim, C. Boyd, J. Paulson, J. Charles and J. Petrie, "Agent-based electronic markets for project Supply Chain", In: Proceedings of the Knowledge-based Electronic Markets, at the AAAI '00 Workshop, July 31, Austin, TX, USA. 2000.

[28] S. D. Pathak, G. Nordstrom and S. Kurokawa, "Modeling of Supply Chain: A multi-agent approach", In: Proceeding on IEEE SMC 2000, CD ROM Reference 00ch37166c, Nashville, TN, October 8, 2000.

[29] M. He and H-F. Leung, "Agents in E-Commerce: State of the art", Knowledge and Information Systems 4 (3), 2002, 257–282.

[30] J. Wu, M. Cobzaruo, M. Ulieru and D.H, Norrie, "SC-Web- CS: Supply Chain web-centric systems", In: Proceedings of the IASTED International Conference on Artificial, Banff, July 24–26,2000, pp. 501–507.

[31] J. Hendler, "Agents and the semantic web", IEEE Intelligent Systems 16 (2), 2001, 30–37.

[32] T. Berners-Lee, "Axioms, architecture and aspirations", In: W3C All-working Group Plenary Meeting, February 28, 2001.

[33] K. Turowski, "Agent-based e-commerce in case of mass customization", International Journal of Production Economics 75, 2002, 69–81.

[34] S. B. Yoo and Y. Kim, "Web-based knowledge management for sharing product data in virtual enterprises", International Journal of Production Economics 75, 2002, 173–183.

[35] T. Kaihara, "Multi-agent based Supply Chain modeling with dynamic environment", International Journal of Production Economics 85, 2003, 263–269.

[36] R. Singh, A.F, Salam, L. Iyer, "Agents in e-Supply Chains", Communications of the ACM 48 (6), 2005, 108–115.

[37] J. Reaidy, P. Massotte, D. Diep, "Comparison of negotiation protocols in dynamic agent-based manufacturing systems", International Journal of Production Economics 99, 2006, 117–130.

[38] N. Guarino, "Formal Ontology and Information Systems", Formal Ontology in Information Systems. IOS Press, 1998.